\documentclass[preprint,3p,11pt,times,twocolumn]{elsarticle}

\usepackage{amssymb}
\usepackage{amsmath}
\usepackage[dvipsnames,table,xcdraw]{xcolor} 
\usepackage{pifont}                          
\usepackage{booktabs}
\usepackage[colorlinks=true]{hyperref}
\usepackage{xurl}
\usepackage{caption}

\journal{xxxxxxxxxx}

\begin{document}

\hypersetup{
    colorlinks=true,
    citecolor=RoyalBlue,
    urlcolor=black,
    linkcolor=RoyalBlue
}

\begin{frontmatter}



\title{HierLight-YOLO: A Hierarchical and Lightweight Object Detection Network for UAV Photography}


\author[label1]{Defan Chen} 
\author[label1]{Yaohua Hu}

\author[label1]{Luchan Zhang\corref{cor1}}

\cortext[cor1]{Corresponding author}
\ead{zhanglc@szu.edu.cn}

\affiliation[label1]{organization={School of Mathematical Sciences},
            addressline={Shenzhen University},
            city={Shenzhen},
            postcode={518060},
            state={Guangdong},
            country={China}}

\begin{abstract}
The real-time detection of small objects in complex scenes, such as the unmanned aerial vehicle (UAV) photography captured by drones, 
has dual challenges of detecting small targets ($<$32×32 pixels) and maintaining real-time efficiency on resource-constrained platforms. 
While YOLO-series detectors have achieved remarkable success in real-time large object detection, they suffer from significantly higher false negative rates for drone-based detection where small objects dominate, compared to large object scenarios. 
This paper proposes HierLight-YOLO, a hierarchical feature fusion and lightweight model that enhances the real-time detection of small objects, based on the YOLOv8 architecture. 
We propose the Hierarchical Extended Path Aggregation Network (HEPAN), a multi-scale feature fusion method through hierarchical
cross-level connections, enhancing the small object detection accuracy. 
HierLight-YOLO includes two innovative lightweight modules: Inverted Residual Depthwise Convolution Block (IRDCB) and Lightweight Downsample (LDown) module, which significantly reduce the model's parameters and computational complexity without sacrificing detection capabilities. 
Small object detection head is designed to further enhance spatial resolution and feature fusion to tackle the tiny object (4×4 pixels) detection. 
Comparison experiments and ablation studies on the VisDrone2019 benchmark demonstrate state-of-the-art performance of HierLight-YOLO:  
the nano-scale HierLight-YOLO-N maintains strong detection capability (35.8\% AP\textsubscript{0.5}) at just 2.2M parameters (26.7\% fewer than YOLOv8-N), proving its suitability for edge deployment;  
the small-scale HierLight-YOLO-S achieves the state-of-the-art accuracy 44.9\% AP\textsubscript{0.5} among S-scale models with only 7.8M parameters (29.7\% fewer than YOLOv8-S); 
the medium-scale variant HierLight-YOLO-M outperforms other M-scale models by 50.2\% AP\textsubscript{0.5} (surpassing YOLOv8-M by 5.6\%), with  competitive 17.9M parameters. 
Ablation studies validate each component’s contribution, with HEPAN improving AP\textsubscript{0.5} by 0.8\% over conventional feature pyramids and IRDCB reducing parameters by 22.1\% without accuracy loss. 
\end{abstract}




\end{frontmatter}



\section{Introduction}
\label{sec:intro}

As drone technology rapidly advances, aerial photography has become an essential tool in various fields, such as disaster monitoring, traffic management, search and rescue, and agricultural oversight. 
Unlike traditional ground-based methods, drone imagery provides a high-altitude perspective, wide coverage, and reduced operational costs.
Real-time, accurate detection is crucial for effective drone deployment across these applications - whether identifying survivors and hazards in disaster response or detecting subtle anomalies for precision crop management. 
However, small object detection in drone imagery presents significant challenges, including complex backgrounds, low-contrast objects, and dynamic environmental conditions, which frequently result in missed detections and false alarms \citep{wu2021deep,tang2023survey}. 
Overcoming these limitations would enable more reliable and efficient drone operations, maximizing their potential in demanding real-world scenarios. 

Existing object detection methods can be broadly categorized into two approaches: two-stage models, which first generate region proposals and then classify and refine them, such as the R-CNN series \citep{girshick2014rich,girshickICCV15fastrcnn,renNIPS15fasterrcnn}, and single-stage models, which perform detection in a single pass by treating object detection as a regression problem, such as the YOLO series  \citep{redmon2016you,redmon2017yolo9000,farhadi2018yolov3,bochkovskiy2020yolov4,li2022yolov6,wang2023yolov7,wang2024yolov9,wang2024yolov10,Jocher_Ultralytics_YOLO_2023}. 
While two-stage methods offer high accuracy through region proposal networks, their computational complexity often hinders real-time performance. In contrast, single-stage models achieve faster inference, though traditionally with some accuracy trade-offs. 
Recent advancements of the YOLO family have bridged critical performance gaps through anchor-free detection and enhanced feature fusion to achieve a speed-accuracy balance for real-time applications.

Significant challenges remain in detecting small objects within complex aerial scenes. 
These limitations persist due to inherent difficulties in capturing discriminative features from low-resolution objects among cluttered backgrounds and scale variations.

Recent studies have pursued several architectural strategies to enhance small object detection. 
Multi-scale feature fusion approaches have evolved from the seminal Feature Pyramid Network (FPN) \citep{lin2017feature}, which established hierarchical feature combination through top-down pathways with lateral connections, to more sophisticated variants like Path Aggregation Network (PANet) \citep{liu2018path} that introduced complementary bottom-up information flow.  
The field further advanced with Bidirectional FPN (BiFPN)'s learnable cross-scale feature weighting, enabling dynamic importance allocation across resolution levels \citep{tan2020efficientdet}. 
Parallel developments in attention mechanisms have introduced content-aware feature modulation, allowing networks to focus computational resources on semantically salient regions \citep{vaswani2017attention}.
Meanwhile, lightweight design paradigms have emerged through architectural innovations such as depthwise separable convolutions in MobileNets \citep{howard2017mobilenets,sandler2018mobilenetv2,howard2019searching}, channel shuffling operations in ShuffleNets \citep{zhang2018shufflenet,ma2018shufflenet}, and neural architecture search-derived scaling principles in EfficientNets \citep{tan2019efficientnet,tan2021efficientnetv2}.  
While these methods perform well in reducing computational overhead for general object detection, they remain insufficient to deal with  extremely small object detection against cluttered backgrounds and varying scales in drone-specific scenarios. 
These persistent challenges highlight the need for specialized solutions in aerial imagery analysis.

To address the limitations of existing models in small object detection of the complex drone-captured scenarios, we propose HierLight-YOLO, an enhanced lightweight architecture based on YOLOv8 that significantly improves small object detection while maintaining computational efficiency for resource-constrained devices. 

The major contributions of this work include:
\begin{itemize}
\item \textbf{Hierarchical Feature Fusion Architecture:} We propose the Hierarchical Extended Path Aggregation Network (HEPAN), which extends conventional PANet through two key innovations: (1) intermediate convolutional blocks with residual connections to strengthen gradient propagation, and (2) bidirectional multi-scale feature fusion that preserves both high-resolution details and rich semantic information. It enhances multi-scale feature fusion through hierarchical cross-level connections, substantially improving small object recognition. 
    
\item \textbf{Lightweight Network Optimization:} Our redesigned Inverted Residual Depthwise Convolution Block (IRDCB) module combines depthwise separable convolutions \citep{chollet2017xception} with an inverted residual structure \citep{sandler2018mobilenetv2}, achieving 22.1\% fewer parameters than the original C2f module while maintaining comparable accuracy. 
    The introduced Lightweight Downsample (LDown) module further reduces parameter count by 11.4\% through optimized channel compression, enabling real-time operation on edge devices with limited computational resources. 
\item \textbf{Small Object Detection Head:} We introduce an additional detection head specifically designed for small objects. By strategically fusing shallow spatial features with deep semantic features through cross-layer connections, this enhancement significantly improves detection recall for objects smaller than 32×32 pixels while maintaining computational efficiency.
\end{itemize}

\begin{figure*}[h]
\includegraphics[width=1\linewidth]{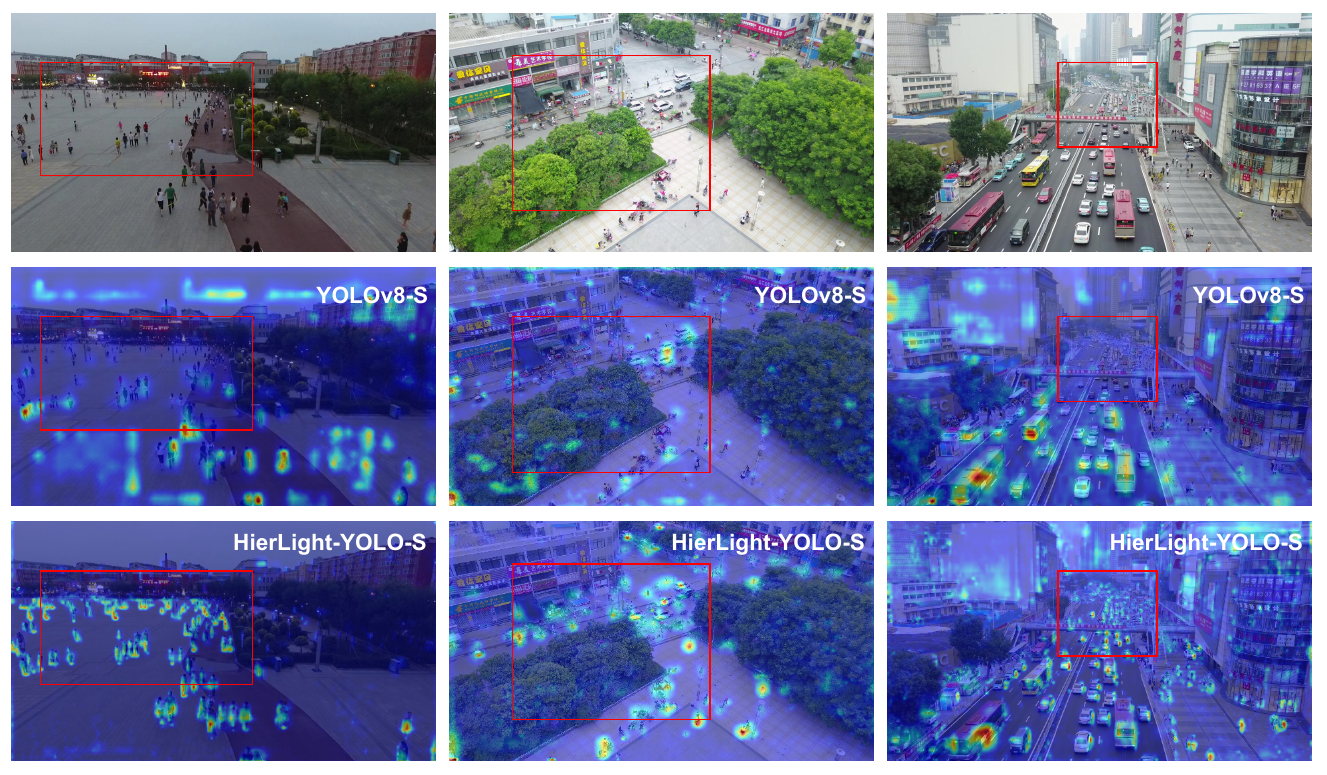}
\caption{Comparison of heat maps between baseline YOLOv8-S model and our proposed HierLight-YOLO-S model. 
The color spectrum (blue → red) indicates relative attention intensity, with red regions denoting high model focus.
Our HierLight-YOLO-S model shows significantly improved attention allocation to small objects and their surroundings compared to the baseline YOLOv8-S.  
The red bounding boxes highlight regions where HierLight-YOLO-S achieves superior small object detection.
\label{fig1}}
\end{figure*}

Through comprehensive evaluation on the VisDrone2019 benchmark \citep{du2019visdrone}, our proposed HierLight-YOLO models demonstrates significant improvements over baseline YOLOv8 models across multiple metrics. 
The proposed architecture achieves a 3.3\% increase in average precision (AP) for small objects ($<$32×32 pixels) while maintaining real-time processing speeds of 133 FPS on an NVIDIA RTX 6000 Ada. 
As visualized in Fig.~\ref{fig1}, our model's class activation heatmaps reveal two critical advantages: (1) stronger thermal responses concentrated on small objects (3-5× higher intensity values compared to baseline YOLOv8), and (2) more precise spatial localization evidenced by tighter heat distribution around object boundaries. 
These visualizations quantitatively confirm that HEPAN’s hierarchical feature fusion and the improved lightweight modules effectively enhance the model’s ability to capture fine-grained features in challenging drone scenarios.

\section{Related Work}
\label{sec:relat}

A fundamental challenge in object detection lies in effectively recognizing objects across varying scales, particularly small objects that often comprise fewer than 0.1\% of an image's pixels. 
These small objects suffer from severe feature degradation during repeated convolution operations in deep convolutional networks, making them prone to being lost in deeper layers. 
The research community has progressively developed increasingly sophisticated multi-scale feature fusion approaches to address this limitation. 
Early detection frameworks such as R-CNN \citep{girshick2014rich} and Fast R-CNN \citep{girshickICCV15fastrcnn} relied on single-scale feature extraction, inherently limiting their multi-scale detection capability. 
While Faster R-CNN \citep{renNIPS15fasterrcnn} introduced the Region Proposal Network (RPN) for candidate generation, it remained constrained by fixed-scale feature processing. 
A breakthrough came with the FPN \citep{lin2017feature}, which established the now-standard paradigm of top-down multi-resolution feature fusion through lateral connections. 
Subsequent innovations have further refined this approach: PANet \citep{liu2018path} augmented FPN with bottom-up path augmentation to strengthen feature flow, while Adaptively Spatial Feature Fusion (ASFF) \citep{liu2019learning} introduced adaptive spatial weighting to dynamically emphasize the most discriminative features across scales. 
The field has also explored automated architecture design with Neural Architecture Search Feature Pyramid Network (NAS-FPN) \citep{ghiasi2019fpn}, though at considerable computational expense. 
Currently, BiFPN \citep{tan2020efficientdet} represents the state-of-the-art through its learnable cross-scale feature weighting mechanism, demonstrating that carefully designed fusion strategies can significantly boost small object detection without prohibitive computational overhead.

Recent advances in lightweight network design have significantly enhanced the deployment of deep learning models in resource-constrained environments. 
The field has evolved through several key innovations: MobileNet \citep{howard2017mobilenets} pioneered the use of depthwise separable convolutions, reducing computational complexity by up to 8-9× compared to standard convolutions while maintaining competitive accuracy. 
ShuffleNet \citep{ma2018shufflenet} advanced this paradigm by introducing channel shuffling operations to enable efficient cross-group information exchange in grouped convolutions. 
A major breakthrough came with EfficientNet \citep{tan2019efficientnet}, which established a principled compound scaling method to jointly optimize network width, depth, and resolution. 
Subsequent architectures like CSPNet \citep{wang2020cspnet} demonstrated that cross-stage partial connections could simultaneously reduce computation by 20\% while enhancing gradient flow. 
GhostNet \citep{han2020ghostnet} further pushed efficiency boundaries through its ghost module, which generates additional feature maps via cheap linear operations rather than expensive convolutions. 
These innovations collectively represent an evolutionary trajectory where each new architecture builds upon and refines the core ideas of its predecessors, progressively improving the trade-off between computational efficiency and model performance for edge deployment scenarios.

Real-time end-to-end object detection has achieved remarkable progress through the dual evolution of YOLO-series architectures and transformer-based approaches. The YOLO family \citep{redmon2016you,redmon2017yolo9000,farhadi2018yolov3,bochkovskiy2020yolov4,li2022yolov6,wang2023yolov7,Jocher_Ultralytics_YOLO_2023,wang2024yolov9} established the dominant paradigm for efficient detection via unified regression frameworks. Recent innovations like YOLOv10 \citep{wang2024yolov10} demonstrate further improvements through introducing transformer-enhanced hybrid encoders that eliminate non-maximum suppression (NMS) bottlenecks while maintaining \textless2ms latency on GPU hardware. RT-DETR \citep{zhao2024detrs} further revolutionized this field by employing CNN backbones with a transformer-based hybrid encoder (featuring scale-aware feature fusion via AIFI and CCFM modules) and decoder, enhanced through uncertainty-minimal query selection. 
Parallel advancements in training efficiency include DEIM \citep{huang2024deim}, which reduces convergence time by 40\% through dense one-to-one matching, and D-FINE \citep{peng2024dfine} that improves occlusion handling by 18\% AP through decoupled position-size prediction. These developments collectively push the boundaries of real-time detection in complex scenarios.
The latest YOLOv12 \citep{tian2025yolov12} integrates area attention mechanism and residual ELAN module to optimize multi-scale perception, and surpasses a number of models through flash attention mechanism optimization. 
These developments collectively push the boundaries of real-time detection in complex scenarios.

\section{Methodology}

\begin{figure}[htb!]
\includegraphics[width=1\linewidth]{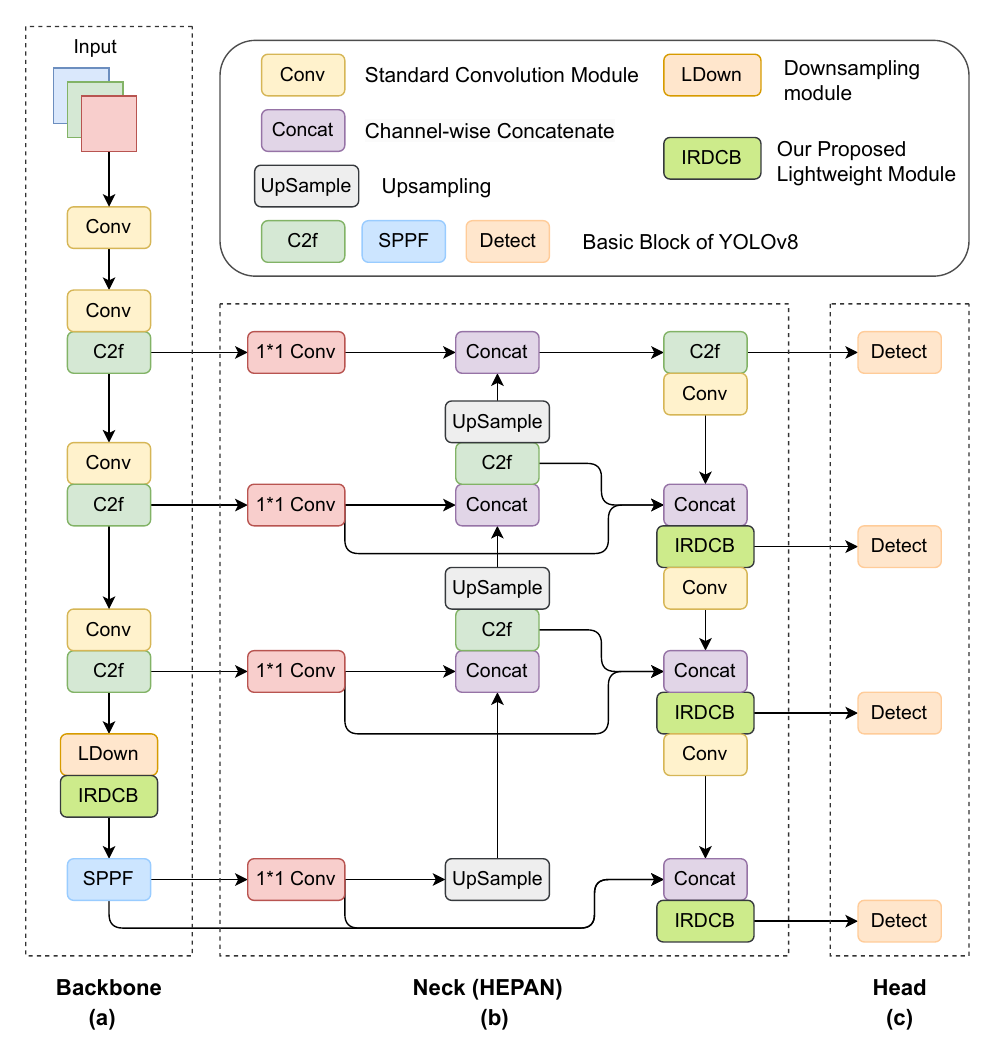}
\caption{HierLight-YOLO architectural overview: (a) Backbone with IRDCB feature extraction block and LDown downsampling module; 
(b) Hierarchical feature fusion via HEPAN neck; (c) Multi-scale detection head.\label{fig4}}
\end{figure}

In this section, we present the details of the network design of our HierLight-YOLO model. 
Our goal is to improve the detection performance of the YOLOv8 model, especially on the challenging task of small object detection in drone imagery. 
We enhance the small object detection by focusing on optimizing the multi-scale feature fusion mechanism and introducing lightweight modules. 
These innovations enable the model to capture the key feature information of small objects while maintaining excellent accuracy in complex and cluttered backgrounds. 
In the following subsections, we will present the core aspects of these enhancements and explore their detailed implementation. 
The overall structure of our enhanced network is demonstrated in Fig.~\ref{fig4}. 
It consists of three main components: the Backbone, the Neck, and the Head. The Backbone includes standard convolution (Conv), C2f, and our lightweight IRDCB modules, responsible for feature extraction and compression. The Neck uses the HEPAN to fuse multi-scale features, enhancing small object detection. The Head contains four detection layers of different scales to ensure that the model can accurately identify objects at multiple scales.

\subsection{Hierarchical Feature Fusion Architecture}

\begin{figure}[htbp!]
\includegraphics[width=1\linewidth]{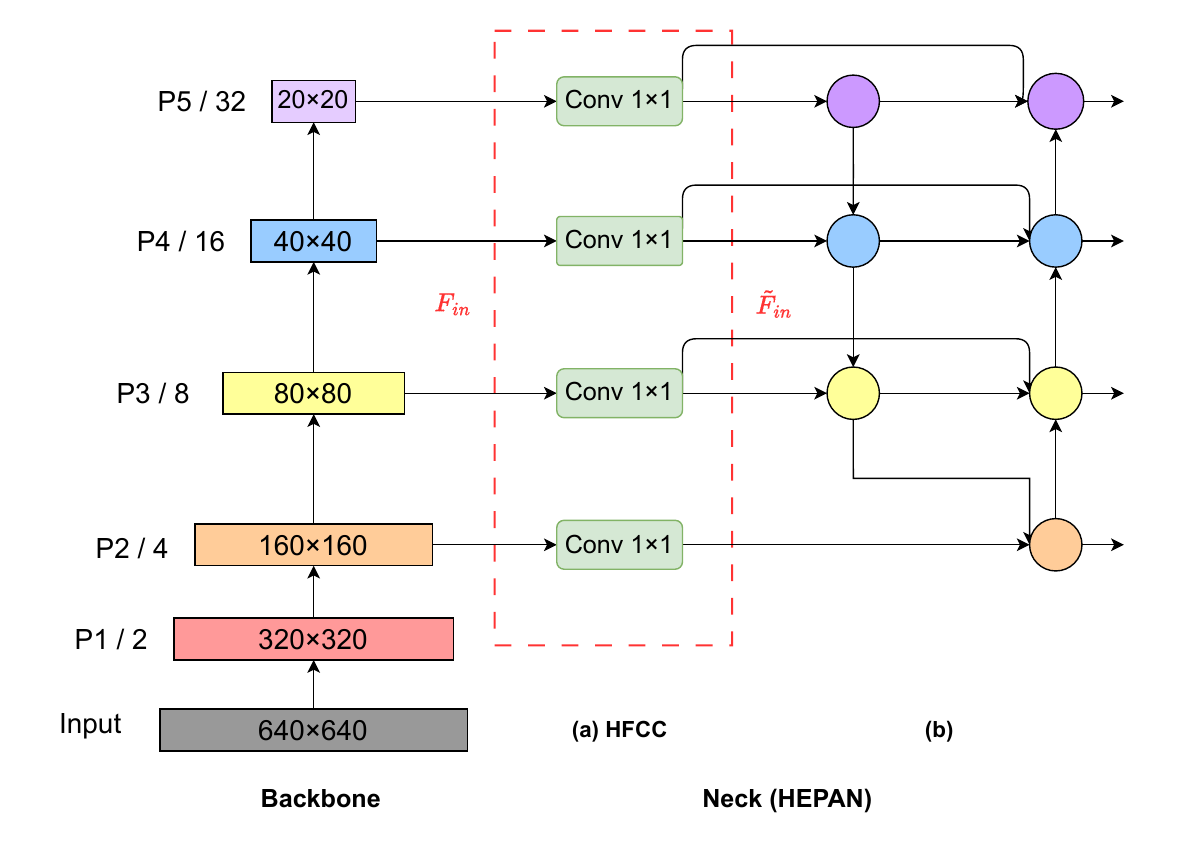}
\caption{HEPAN architecture: 
(a) HFCC compresses input features ${F}_{\text{in}}$ via 1×1 convolutions;  
(b) Bidirectional feature fusion with dense skip connections.
\label{fig2}}
\end{figure}

We propose the Hierarchical Extended Path Aggregation Network (HEPAN) to optimize feature fusion for small object detection, as shown in Fig.~\ref{fig2}.
By introducing additional 1×1 convolutional layers in the neck network, HEPAN enhances feature extraction and representation capabilities, while dense residual connections are included to improve gradient flow stability. 
Compared to traditional PANet and BiFPN structures, HEPAN achieves superior detection accuracy, particularly for small objects in complex backgrounds, through more detailed feature connections and efficient information flow.

\subsubsection*{Hierarchical Feature Channel Compression (HFCC)}

Our hierarchical channel compression design, implemented through 1×1 convolutional layers at backbone outputs (P2-P5), is theoretically grounded in the Information Bottleneck Principle \citep{7133169,shwartz2017opening,icml2023kzxinfodl},  
which states that effective feature representation requires compressing redundant information while preserving task-relevant semantics. For each hierarchical input feature ${F}_{\text{in}}$, the channel compression operation performed by the 1×1 convolution to the compressed features $\tilde{F}_{\text{in}}$ as
\begin{equation}
    \tilde{F}_{\text{in}} =\text{Conv}^{({\text{in}})}_{1\times1}(F_{\text{in}}).
\end{equation}
This operation serves two key purposes: (1) it maximizes the mutual information $I(Y;\tilde{F}_{\text{in}})$ between the compressed features $\tilde{F}_{\text{in}}$ and the detection object $Y$, while (2) minimizing $I(\tilde{F}_{\text{in}};F_{\text{in}})$ to filter out irrelevant background noise and textures. 
Through strategic channel dimension reduction (for example, compressing the P5 features from 1024 to 512 channels), the model maintains efficient computation while prioritizing the most important spatial and semantic information needed for accurate multi-scale object detection, particularly beneficial for lightweight network architectures.

\subsubsection*{Cross-Layer Dense Skip Connections}

Building upon dense connection theory \citep{huang2017densely}, our architecture employs cross-layer skip connections to enhance gradient propagation efficiency in deep networks. For shallow features $F_l$, gradients flow to deeper fusion nodes $F_{l+j}, j=1,\cdots,m$ through
\begin{equation}
    \frac{\partial \mathcal{L}}{\partial F_l} = \sum_{j=1}^{m} \frac{\partial \mathcal{L}}{\partial F_{l+j}}  \frac{\partial F_{l+j}}{\partial F_l}.
\end{equation}
This mechanism preserves high-resolution details in deep feature computation while mitigating gradient vanishing, particularly benefiting small object parameter updates. 

We extend the bidirectional flow philosophy from PANet through a cross-layer aggregation structure that combines P2’s fine-grained details and P5’s global context, resolving spatial-semantic disparities to ensure robust scale variation handling.

The combined operation of channel compression and skip connections achieves optimal efficiency-representation balance. 
The 1×1 convolutions serve as information filters, while cross-layer links create dense feedback loops that emphasize salient multi-scale features - a critical capability for small-object detection in complex environments.

\subsection{Improved Lightweight Module}

In the small object detection tasks, the computational complexity and parameter size of the model have a crucial impact on real-time performance and efficiency, especially when deployed on resource-constrained devices. 
Consequently, lightweight model design is essential for enhancing detection speed and reducing power consumption. 
We optimize the model architecture to reduce redundant computations while maintaining detection accuracy. 
By designing a novel feature extraction module, we significantly decrease both the number of model parameters and computational costs without compromising performance. 

\subsubsection*{Inverted Residual Depthwise Convolution Block (IRDCB)}
The IRDCB module reconstructs feature fusion through an efficient architecture combining depthwise separable convolutions and channel rearrangement, achieving significant computational cost reduction while preserving feature representation quality. 
Building upon MobileNetV2's inverted residual structure \citep{sandler2018mobilenetv2}, our module design specifically addresses multi-scale feature requirements in object detection. 

\begin{figure*}[htb!]
    \centering
    \includegraphics[width=1.0\linewidth]{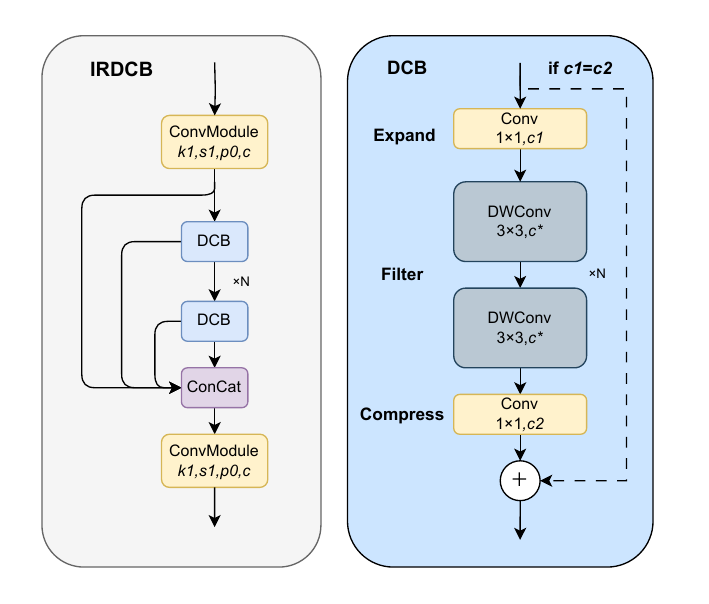}
    \caption{IRDCB module structure. 
    }
    \label{fig3}
\end{figure*}

As demonstrated in Fig.~\ref{fig3},  our IRDCB module implements an efficient feature transformation $\mathcal{F}$ for the input feature $\mathbf{x}_{\text{in}}$ through three sequential phases with channel dimensions $c_1$ (input) and $c_2$ (output):
\begin{equation}
\mathcal{F}(\mathbf{x}_{\text{in}}) = \underbrace{\text{Conv}_{1\times1}^{\downarrow}}_{\text{Compress}} \circ 
\underbrace{\text{DWConv}_{3\times3} \circ \text{DWConv}_{3\times3}}_{\text{Filter}} \circ 
\underbrace{\text{Conv}_{1\times1}^{\uparrow}}_{\text{Expand}}(\mathbf{x}_{\text{in}}).
\end{equation}

\begin{itemize}
    \item \textbf{Expand Phase}. An 1×1 convolution projects input channels $c_1$ to expanded dimension $c^* = \lfloor c_1  t \rfloor$ (expansion factor $t \geq 1$). This operation achieves dual objectives: (1) preventing the feature degradation through richer channel space, which enhances fine-grained detail capture in subsequent depthwise convolutions; (2) improving the computation efficiency, as the 1×1 convolution cost ($c_1c^*HW$) substantially lower than standard convolution ($k^2c_1c^*HW$ for kernel size $k$). 
    
    \item \textbf{Filter Phase}. The $c^* HW$ expanded features are input into dual-layer depthwise convolution (DWConv) with the group number $c^*$ in each layer, implementing channel-wise spatial filtering. 
        The computation complexity of our channel spatial filtering DWConv is $c^* k^2 HW$, which achieves a significant computaitonal efficiency improvement from the standard convolution with complexity $c^{*2} k^2 HW$.  
        Additionally, it enhances nonlinear representation through two independent spatial filtering operations using 3×3 kernels with stride 1 and padding 1, effectively capturing complex local patterns such as edges and textures while mitigating gradient vanishing in deep networks.
    
    \item \textbf{Compress Phase}. The filtered features $c^*HW$ are compressed back to the object channel number $c_2$ via a 1×1 convolution, which simultaneously performs feature selection and dimensionality alignment. 
        This operation functions as an efficient channel attention mechanism, where learnable kernels selectively preserve shallow spatial detail channels critical for small object detection while suppressing redundant background textures.    
        The transformation ensures output compatibility ($c_2$ channels) for subsequent fusion operations with other backbone features, maintaining both representational quality and architectural flexibility.

    \item \textbf{Residual Connection}. The module implements a conditional residual path governed by channel dimensions,
    \begin{equation}
    \mathbf{x}_{\text{out}} = 
    \begin{cases}
    \mathbf{x}_{\text{in}} + \mathcal{F}(\mathbf{x}_{\text{in}}), & \text{if } c_1 = c_2 \\
    \mathcal{F}(\mathbf{x}_{\text{in}}), & \text{otherwise}
    \end{cases}
    \label{rs}
    \end{equation}
    where $\mathbf{x}_{\text{in}}$ is the input feature with dimension $c_1$, and $\mathbf{x}_{\text{out}}$ is the output feature with dimension $c_2$.
    This design provides dual-mode operation: identity mapping with skip connection when $c_1=c_2$ to preserve gradient flow in deep networks; and pure transformation when channel dimensions mismatch (e.g., during cross-level feature fusion), ensuring optimal feature adaptation throughout the network hierarchy.  
    
\end{itemize}

\subsubsection*{Lightweight Downsample (LDown) Module}
The LDown module is designed to efficiently reduce both spatial dimensions and channel capacity of feature maps, optimizing computational efficiency while preserving critical multi-scale information. 
It consists of two cascaded convolutional layers: a spatial-refining layer and a channel-compressing layer, with configurable kernel size $k$ and stride $s$ to control downsampling strength,  
\begin{equation}
    F_{\text{out}} = \text{Conv}_{1 \times 1}^{(2)} \circ \text{Conv}_{k \times k, s}^{(1)}(F_{\text{in}}),
\end{equation}
where $F_{\text{in}}$ is the input feature, and $F_{\text{out}}$ is the output feature.
The first layer $ \text{Conv}_{k \times k, s}^{(1)}$ applies a $k \times k$ convolution with stride s and group size $g = c_1$ (equivalent to a depthwise convolution). 
With stride $s$, the $k \times k$ kernel reduces the spatial resolution of the input feature map (e.g., $H \times W \to \lfloor H/s \rfloor \times \lfloor W/s \rfloor$), enabling the model to capture coarser-grained context at lower computational cost.  
The second layer $\text{Conv}_{1 \times 1}^{(2)}$ follows with a 1×1 convolution to transform the channel dimension from $c_1$ to $c_2$. The 1×1 kernel operates on each spatial position independently, mixing channel information without altering spatial resolution. Its lower complexity ensures subsequent layers process fewer but more informative features.

\subsection{Small Object Detection Head}

To address the challenge of detecting objects smaller than 32×32 pixels in UAV images, we introduce a dedicated small object detection head to enhance spatial resolution and feature fusion. This method extends the traditional P3-P5 hierarchy with a high-resolution (160×160) layer.
The small object detection layer operates through a two-stage feature transformation framework, defined as a composition of upsampling and fusion operations. 

The upsampling operation $\phi_s$ is designed as the nearest-neighbor upsampling with a scaling factor $s$, 
\begin{equation}
    F_{\text{upsamp}}(c, x, y)=\phi_s(F) = F\left(c, \left\lfloor \frac{x}{s} \right\rfloor, \left\lfloor \frac{y}{s} \right\rfloor \right),
\end{equation}

where for an input feature map $F \in \mathbb{R}^{C \times H \times W}$, the upscaled output $\phi_s(F) \in \mathbb{R}^{C \times sH \times sW} $is computed by replicating each pixel $(c, i, j)$ in $F$ to an $s \times s$ block in the output.

In the first stage of our feature upsampling, we employ upsampling operation $\phi_2$ in which the nearest-neighbor interpolation transform the $80 \times 80$ resolution features $F_{\text{P3}} \in \mathbb{R}^{C \times 80 \times 80}$ from the neck network's P3 layer into a $160 \times 160$ feature map. 

It ensures that each $2 \times 2$ block in the upsampled feature map corresponds to a single pixel in the original feature map, preserving the original pixel values without introducing any interpolation artifacts. This upsampling step elevates the spatial resolution of the feature map, enabling better alignment with higher-layer features in subsequent fusion stages and retaining fine-grained details critical for small object detection.

In the second stage, we obtain a merged concatenated feature by using the channel-wise concatenation to combine the upsampled features $\phi_2(F_{\text{P3}}) \in \mathbb{R}^{C_1 \times 160 \times 160}$ with shallow P2 features $F_{\text{P2}} \in \mathbb{R}^{C_2 \times 160 \times 160}$ from the backbone network, 
\begin{equation}
    F_{\text{merge}} = \text{Concat}(\phi_2(F_{\text{P3}}), F_{\text{P2}}).
    \label{eq:concat2}
\end{equation} 
This fusion process concatenates feature maps along the channel dimension, preserving both spatial information and channel-wise semantics. 

This approach effectively merges the spatial detail features from the shallow P2 layer, which capture edge and texture information, with the semantic context features from the deeper neck layers.

The resultant 160×160 feature map offers a spatial sampling density fourfold higher than the standard P3 layer. The increased sampling density endows the model with several advantages. It enables the model to detect objects as small as 4×4 pixels, corresponding to 1/40 of a feature map cell. This finer granularity reduces the likelihood of false negatives that often occur due to missed annotations of small objects. 

\section{Experiments and Analysis}

\subsection{Dataset}

We conduct our experiments using the VisDrone2019 dataset \citep{du2019visdrone}, an aerial imagery benchmark developed by Tianjin University. This dataset contains 10,209 annotated images (2000×1500 to 480×360 resolution) captured from drone perspectives, with official splits for training (6,471 images), validation (548 images), test (1610 images), and competition sets (1580 images). 

Compared to ground-level datasets like MS COCO \citep{lin2014microsoft} and Pascal VOC2012 \citep{everingham2015pascal}, VisDrone presents unique challenges including distinct camera angles (nadir and oblique views), diverse lighting/weather conditions,  and complex backgrounds (urban, park, and school environments). 
The dataset annotates 10 object categories critical for drone applications, including pedestrians, cars, bicycles, and traffic devices, making it particularly suitable for evaluating small object detection in real-world scenarios.

\subsection{Experimental Environment}

Our experiments were conducted on an Ubuntu 24.04.1 system with Python 3.10.14, PyTorch 2.3.1, and CUDA 11.8, utilizing an NVIDIA RTX 6000 Ada GPU. Based on the Ultralytics YOLOv8 framework, we trained models for 600 epochs with 640×640 input resolution, employing the Adam optimizer (lr=0.001, momentum=0.9) and early stopping (patience=20). 
Evaluation followed COCO metrics including AP$_{0.5:0.95}$ (IoU thresholds 0.50–0.95), AP$_{50}$ (IoU thresholds 0.50),
ensuring comprehensive performance assessment under standardized conditions.

\subsection{Comparison with State-of-the-arts}

\begin{table*}
\centering
\captionsetup{font=normalsize}
\caption{Experimental results compared with other models in VisDrone2019-val.\label{table1}}
\begin{tabular}{ccccccccc}
\toprule
\textbf{Model} & \textbf{\boldmath AP$_{0.5}(\%)$} & \textbf{\boldmath AP$_{0.5:0.95}(\%)$} & \textbf{Params(M)} & \textbf{FLOPs(G)} \\
\midrule
D-FINE-HGNETv2-N \citep{peng2024dfine} & \textbf{36.2} & 20.1 & 3.7 & 7.1 \\
DEIM-HGNETv2-N \citep{huang2024deim} & 36.1 & 20.1 & 3.7 & 7.1 \\
YOLOv8-N \citep{Jocher_Ultralytics_YOLO_2023} & 33.4 & 19.4 & 3.0 & 8.1 \\
YOLOv8-N-P2 \citep{Jocher_Ultralytics_YOLO_2023} & 35.3 & 20.8 & 2.9 & 12.2 \\
YOLOv9-T \citep{wang2024yolov9} & 33.7 & 19.5 & \textbf{2.0} & 7.6 \\
YOLOv10-N \citep{wang2024yolov10} & 32.1 & 18.6 & 2.7 & 8.2 \\
YOLOv11-N \citep{Jocher_Ultralytics_YOLO_2023} & 32.0 & 18.6 & 2.6 & \textbf{6.3} \\
YOLOv12-N \citep{tian2025yolov12} & 33.5 & 19.6 & 2.6 & \textbf{6.3} \\
\rowcolor[gray]{0.8}
\textbf{HierLight-YOLO-N (Ours)} & 35.8 & \textbf{21.0} & 2.2 & 11.7 \\
\midrule
DEIM-HGNETv2-S \citep{huang2024deim} & 44.8 & 26.8 & 10.2 & 25.0 \\
HIC-YOLOv5 \citep{tang2024hic} & 44.3 & 26.0 & 9.3 & 30.9 \\
TPH-YOLOv5-S \citep{zhu2021tph} & 43.1 & 26.3 & 7.5 & 36.7 \\
YOLOv8-S \citep{Jocher_Ultralytics_YOLO_2023} & 40.9 & 24.3 & 11.1 & 28.5 \\
YOLOv8-S-P2 \citep{Jocher_Ultralytics_YOLO_2023} & 44.1 & 26.5 & 10.6 & 36.7 \\
YOLOv9-S \citep{wang2024yolov9} & 40.0 & 23.7 & \textbf{7.2} & 26.7 \\
YOLOv10-S \citep{wang2024yolov10} & 39.1 & 23.1 & 8.0 & 24.5 \\
YOLOv11-S \citep{Jocher_Ultralytics_YOLO_2023} & 39.4 & 23.5 & 9.4 & 21.3 \\
YOLOv12-S \citep{tian2025yolov12} & 40.2 & 24.2 & 9.2 & \textbf{21.2} \\
\rowcolor[gray]{0.8}
\textbf{HierLight-YOLO-S (Ours)} & \textbf{44.9} & \textbf{27.3} & 7.8 & 33.7 \\
\midrule
RT-DETRv2-S \citep{lv2024rtdetrv2improvedbaselinebagoffreebies} & 46.0 & 27.5 & 19.9 & 59.9 \\
RT-DETRv2-M* \citep{lv2024rtdetrv2improvedbaselinebagoffreebies} & 48.2 & 28.6 & 31.1 & 91.9 \\
DEIM-RTDETRv2-R18 \citep{huang2024deim} & 46.9 & 28.0 & 19.9 & 59.9 \\
DEIM-HGNETv2-M \citep{huang2024deim} & 50.1 & 30.4 & 19.2 & 56.4 \\
UAV-DETR-EV2 \citep{zhang2025uav} & 47.0 & 28.2 & \textbf{13.0} & \textbf{42.9} \\
UAV-DETR-R18 \citep{zhang2025uav} & 49.4 & 30.4 & 21.3 & 72.5 \\
YOLOv8-M \citep{Jocher_Ultralytics_YOLO_2023} & 44.6 & 27.3 & 25.8 & 78.7 \\
YOLOv8-M-P2 \citep{Jocher_Ultralytics_YOLO_2023} & 49.2 & 30.3 & 25.0 & 98.0 \\
YOLOv9-M \citep{wang2024yolov9} & 45.3 & 27.7 & 20.0 & 76.5 \\
YOLOv10-M \citep{wang2024yolov10} & 44.7 & 27.4 & 16.5 & 63.5 \\
YOLOv11-M \citep{Jocher_Ultralytics_YOLO_2023} & 45.3 & 27.8 & 20.0 & 67.7 \\
YOLOv12-M \citep{tian2025yolov12} & 44.9 & 27.3 & 20.1 & 67.2 \\
\rowcolor[gray]{0.8}
\textbf{HierLight-YOLO-M (Ours)} & \textbf{50.2} & \textbf{31.0} & 17.9 & 88.2 \\
\bottomrule
\end{tabular}
\end{table*}

We present a performance comparison between HierLight-YOLO and other available popular object detection models,  as demonstrated in Table~\ref{table1}. 
HierLight-YOLO achieves state-of-the-art object detection accuracy performance for three different scale variants (HierLight-YOLO-N/S/M), while maintaining competitive computational efficiency. 

For nano-scale models, HierLight-YOLO-N attains 35.8\% AP\textsubscript{0.5} and 21.0\% AP\textsubscript{0.5:0.95}, outperforming the YOLO variants YOLOv8-N \citep{Jocher_Ultralytics_YOLO_2023}, YOLOv8-N-P2 \citep{Jocher_Ultralytics_YOLO_2023}, YOLOv9-T \citep{wang2024yolov9}, YOLOv10-N \citep{wang2024yolov10}, YOLOv11-N \citep{Jocher_Ultralytics_YOLO_2023}, and YOLOv12-N \citep{tian2025yolov12} by 2.4\%, 0.5\%, 2.1\%, 3.7\%, 3.8\%, 2.3\% in AP\textsubscript{0.5}, 
while having fewer parameters (2.2M) than most counterparts.

For small-scale models, HierLight-YOLO-S achieves the highest detection accuracy 44.9\% AP\textsubscript{0.5} and 27.3\% AP\textsubscript{0.5:0.95} among all compared S-size models.
It surpasses DEIM-HGNETv2-S \citep{huang2024deim}, HIC-YOLOv5 \citep{tang2024hic}, TPH-YOLOv5-S \citep{zhu2021tph}, YOLOv8-S \citep{Jocher_Ultralytics_YOLO_2023}, YOLOv8-S-P2 \citep{Jocher_Ultralytics_YOLO_2023}, YOLOv9-S \citep{wang2024yolov9}, YOLOv10-S \citep{wang2024yolov10}, YOLOv11-S \citep{Jocher_Ultralytics_YOLO_2023}, and YOLOv12-S \citep{tian2025yolov12} 
by 0.1\%, 0.6\%, 1.8\%, 4.0\%, 0.8\%, 4.9\%, 5.8\%, 5.5\%, 4.7\% in AP\textsubscript{0.5}, 
while maintaining comparable or even fewer computations and parameters (7.8M). 

For medium-scale models, HierLight-YOLO-M attains 50.2\% AP\textsubscript{0.5} and 31.0\% AP\textsubscript{0.5:0.95}, outperforming the compared models including RT-DETRv2-S \citep{lv2024rtdetrv2improvedbaselinebagoffreebies}, RT-DETRv2-M* \citep{lv2024rtdetrv2improvedbaselinebagoffreebies}, DEIM-RTDETRv2-R18 \citep{huang2024deim}, DEIM-HGNETv2-M \citep{huang2024deim}, UAV-DETR-EV2 \citep{zhang2025uav}, UAV-DETR-R18 \citep{zhang2025uav}, YOLOv8-M \citep{Jocher_Ultralytics_YOLO_2023}, YOLOv8-M-P2 \citep{Jocher_Ultralytics_YOLO_2023}, YOLOv9-M \citep{wang2024yolov9}, YOLOv10-M \citep{wang2024yolov10}, YOLOv11-M \citep{Jocher_Ultralytics_YOLO_2023}, and YOLOv12-M \citep{tian2025yolov12} 
by 4.2\%, 2.0\%, 3.3\%, 0.1\%, 3.2\%, 0.8\%, 5.6\%, 1.0\%, 4.9\%, 5.5\%, 4.9\%, and 5.3\% respectively in AP\textsubscript{0.5}, 
while maintaining competitive parameters (17.9M) relative to its counterparts.

These results collectively highlight HierLight-YOLO's exceptional balance between detection accuracy and computational efficiency. 
Compared to the widely-used YOLO series, our model achieves excellent improvements in AP\textsubscript{0.5} across different scales with comparable or even lower computational budgets. This advantage becomes crucial for drone-based applications where both detection accuracy and real-time processing are critical requirements.

\subsection{Ablation Study}
To validate the effectiveness of our proposed HierLight-YOLO network, we conducted comprehensive ablation studies on the VisDrone-2019 validation set. 
Using YOLOv8s as our baseline model, we incrementally incorporated each proposed improvement to evaluate their individual contributions to detection performance. Table~\ref{table2} presents the detailed ablation results.


\begin{table*}[htbp!]%
\centering  
\captionsetup{font=normalsize}
\caption{Ablation experiment result in VisDrone2019-val.\label{table2}}
\resizebox{\textwidth}{!}{
\begin{tabular}{ccccccccc}  
\toprule
\textbf{Baseline} & \textbf{P2} & \textbf{HEPAN} & \textbf{IRDCB} & \textbf{LDown} & \textbf{\boldmath AP$_{0.5}(\%)$} & \textbf{\boldmath AP$_{0.5:0.95}(\%)$} & \textbf{Params(M)} & \textbf{GFLOPs} \\
\midrule
\ding{51} &  &  &  &  & 43.0 & 26.0 & 11.1 & \textbf{28.5} \\
\ding{51} & \ding{51} &  &  &  & 46.2 & 28.3 & 10.6 & 36.7 \\
\ding{51} & \ding{51} & \ding{51} &  &  & \textbf{47.2} & 29.1 & 11.3 & 38.1 \\
\ding{51} & \ding{51} & \ding{51} & \ding{51} &  & 47.1 & 29.0 & 8.8 & 34.5 \\
\rowcolor[gray]{0.8}
\ding{51} & \ding{51} & \ding{51} & \ding{51} & \ding{51} & \textbf{47.2} & \textbf{29.2} & \textbf{7.8} & 33.7 \\
\bottomrule
\end{tabular}%
} 
\end{table*}

Our experiments systematically evaluated four key components: (1) the P2 layer small object detection head, (2) HEPAN architecture, (3) IRDCB module, and (4) LDown downsampling. 

The baseline YOLOv8s achieves 40.9\% AP$_{0.5}$  and 24.3\% AP$_{0.5:0.95}$ with 11.1M parameters and 28.5G FLOPs.
The addition of the P2 detection head boosts performance to 44.1\% AP$_{0.5}$ (+3.2\%) and 26.5\% AP$_{0.5:0.95}$ (+2.2\%), 
while reducing parameters to 10.6M and increasing computation to 36.7G FLOPs. 
This confirms the P2 layer's effectiveness for small object detection despite the computational overhead. 
Incorporating HEPAN further improves the detection accuracy to 44.9\% AP$_{0.5}$ (+0.8\%) and 27.2\% AP$_{0.5:0.95}$ (+0.7\%), demonstrating its benefits for high-IoU detection. 
Subsequent replacement with IRDCB reduces the model size to 8.8M parameters and 34.5G FLOPs while maintaining comparable accuracy.  
The complete model featuring LDown downsampling achieves optimal efficiency with 44.9\% AP$_{0.5}$, 27.3\% AP$_{0.5:0.95}$, 7.8M parameters, and 33.7G FLOPs, demonstrating the advantages of our integrated design for balancing accuracy and efficiency.

\subsection{Component-level Performance Comparison}

\textbf{Network Structure}. 
Table~\ref{table3} compares the performances of three network structures (PANet, BiFPN, and HEPAN) on VisDrone2019-val. 
HEPAN achieves the best detection performance (44.9\% AP$_{0.5}$) with 11.3M parameters and 38.1G FLOPs, outperforming both PANet (44.1\%) and BiFPN (44.3\%). 
While HEPAN requires slightly higher computation, its accuracy improvement is significant.

\begin{table*}[htbp!]
\centering
\captionsetup{font=normalsize}
\caption{Performance comparison of network structures in VisDrone2019-val.\label{table3}}
\begin{tabular}{ccccccccc}
\toprule
\textbf{Structure} & \textbf{\boldmath AP$_{0.5}(\%)$} & \textbf{Params(M)} & \textbf{FLOPs(G)} \\
\midrule
PANet & 44.1 & 10.6 & 36.7 \\
BiFPN & 44.3 & 10.7 & 37.1 \\
\rowcolor[gray]{0.8}
HEPAN & 44.9 & 11.3 & 38.1 \\
\bottomrule
\end{tabular}
\end{table*}
\noindent\textbf{Module}.
Table~\ref{table4} compares C3, C2f, and IRDCB modules. 
IRDCB performs remarkable reduced parameters 8.8M (22.1\% fewer than C2f module), and 34.5G FLOPs (9.4\% fewer than C2f module), 
with accuracy 44.8\% AP$_{0.5}$ which is slightly 0.1\%  lower than C2f module, demonstrating a good balance between performance and efficiency.

\begin{table*}[htbp!]
\centering
\captionsetup{font=normalsize}
\caption{Performance comparison of modules in VisDrone2019-val.\label{table4}}
\begin{tabular}{ccccccccc}
\toprule
\textbf{Module} & \textbf{\boldmath AP$_{0.5}(\%)$} & \textbf{Params(M)} & \textbf{FLOPs(G)} \\
\midrule
C3 & 44.1 & 9.8 & 36.0 \\
C2f & 44.9 & 11.3 & 38.1 \\
\rowcolor[gray]{0.8}
IRDCB & 44.8 & 8.8 & 34.5 \\
\bottomrule
\end{tabular}
\end{table*}

\noindent\textbf{Expansion Factor}. 
Table~\ref{table5} shows the comparison of the model performance with different expansion factors. 
It can be observed that the expansion factor combination n=2,t=2 achieves the best performance 44.9\% AP$_{0.5}$ with 8.8M parameters, 34.5G FLOPs, outperforming all other factors. 
Increasing t from 2 to 4 (with fixed n=1) provided diminishing returns (+0.6\% AP but +6.8\% parameters). 
Notably, for n=2, the increase of t from 2 to 4 leads to performance degradation with increased parameters, indicating excessive expansion disrupts model optimization.

\begin{table*}[htbp!]
\centering
\captionsetup{font=normalsize}
\caption{Performance comparison of expansion factors (n,t) in IRDCB module.\label{table5}}
\begin{tabular}{ccccccccc}
\toprule
\textbf{Factor} & \textbf{\boldmath AP$_{0.5}(\%)$} & \textbf{Params(M)} & \textbf{FLOPs(G)} \\
\midrule
n=1, t=2 & 44.0 & 8.8 & 34.5 \\
n=1, t=4 & 44.6 & 9.4 & 35.4 \\
\rowcolor[gray]{0.8}
n=2, t=2 & 44.9 & 8.8 & 34.5 \\
n=2, t=3 & 44.5 & 9.1 & 35.0 \\
n=2, t=4 & 44.4 & 9.4 & 35.4 \\
\bottomrule
\end{tabular}
\end{table*}

\section{Conclusions and Discussions}

In this paper, we introduces HierLight-YOLO, an efficient object detection framework specifically designed for drone applications. Through systematic architectural innovations, our model achieves superior performance on small object detection while maintaining computational efficiency suitable for edge deployment. The key achievements, verified against our experimental results, demonstrate significant improvements over existing approaches.

The proposed HEPAN structure enhances multi-scale feature fusion, achieving a 0.8\% increase in AP\textsubscript{0.5} (44.9\% vs 44.1\%) compared to standard PANet architectures, as shown in Table~\ref{table3}. Combined with our novel IRDCB module, the model reduces parameters by 22.1\% (from 11.3M to 8.8M) while preserving detection accuracy. 
The complete HierLight-YOLO-S configuration achieves 44.9\% AP\textsubscript{0.5} with only 7.8M parameters, representing a 29.7\% reduction compared to the baseline YOLOv8-S model.

Our experiments on the VisDrone2019 dataset validate the model's effectiveness across different scales. The nano variant (HierLight-YOLO-N) attains 35.8\% AP\textsubscript{0.5} with 2.2M parameters, outperforming comparable YOLO variants by 2.1-3.8\%. 
The small-scale HierLight-YOLO-S achieves the state-of-the-art accuracy 44.9\% AP\textsubscript{0.5} among S-scale models with only 7.8M parameters (29.7\% fewer than YOLOv8-S). 
The medium-scale HierLight-YOLO-M establishes new state-of-the-art results with 50.2\% AP\textsubscript{0.5}, surpassing the medium-scale variants of YOLO series and RT-DETR series while maintaining competitive parameters (17.9M).

The practical implications of these improvements are particularly valuable for real-world drone operations. Our model's balanced performance across accuracy and efficiency metrics demonstrates its suitability for resource-constrained deployment. The architectural innovations specifically address drone-specific challenges including small objects ($<$32×32 pixels) and complex backgrounds.

Future research directions will focus on three key areas: (1) extending the framework to multi-spectral imaging for all-weather operation, (2) developing adaptive compression techniques for dynamic resource constraints, and (3) incorporating 3D spatial awareness for improved object localization. These advancements will further enhance the model's applicability to evolving drone vision challenges while maintaining its core advantages of efficiency and accuracy.

\section*{Acknowledgments}
This work is supported by the National Natural Science Foundation of China 12201423 (L.C. Zhang), 12222112, 12426311, 32170655 (Y.H. Hu), Shenzhen Science $\&$ Technology Program \url{JCYJ20241202124209011} (L.C. Zhang and Y.H. Hu) and \url{RCYX20231211090222026} (L.C. Zhang), \url{RCJC20221008092753082} (Y.H. Hu), Project of Educational Commission of Guangdong Province 2023ZDZX1017 (Y.H. Hu), Research Team Cultivation Program of Shenzhen University 2023QNT011 (L.C. Zhang and Y.H. Hu).





 \bibliographystyle{elsarticle-num} 
 \bibliography{main}







\end{document}